%% file: kdd19-sigconf.tex
\documentclass[sigconf]{acmart}
\usepackage{balance}
\usepackage{multirow}
\usepackage{enumitem}
\newcommand{\demonet}{{\emph {DEMO-Net}}}
\def\BibTeX{{\rm B\kern-.05em{\sc i\kern-.025em b}\kern-.08emT\kern-.1667em\lower.7ex\hbox{E}\kern-.125emX}}
    
\copyrightyear{2019}
\acmYear{2019}
\setcopyright{acmcopyright}
\acmConference[KDD '19] {The 25th ACM SIGKDD Conference on Knowledge Discovery and Data Mining}{August 4--8, 2019}{Anchorage, AK, USA}
\acmBooktitle{The 25th ACM SIGKDD Conference on Knowledge Discovery and Data Mining (KDD'19), June 22--24, 2019, Anchorage, AK, USA}
\acmPrice{15.00}
\acmDOI{10.1145/3292500.3330950}
\acmISBN{978-1-4503-6201-6/19/08}

\begin{document}

% The "title" command has an optional parameter, allowing the author to define a "short title" to be used in page headers.
\title{\demonet: Degree-specific Graph Neural Networks for\\ Node and Graph Classification}

% The "author" command and its associated commands are used to define the authors and their affiliations.
% Of note is the shared affiliation of the first two authors, and the "authornote" and "authornotemark" commands
% used to denote shared contribution to the research.
\author{Jun Wu}
% \authornote{Dr.~Trovato insisted his name be first.}
% \orcid{1234-5678-9012}
\affiliation{%
  \institution{Arizona State University}
%   \streetaddress{P.O. Box 1212}
%   \city{Tempe}
%   \state{Arizona}
%   \postcode{43017-6221}
}
\email{junwu6@asu.edu}

\author{Jingrui He}
% \authornote{The secretary disavows any knowledge of this author's actions.}
\affiliation{%
  \institution{Arizona State University}
%   \streetaddress{P.O. Box 1212}
%   \city{Tempe}
%   \state{Arizona}
%   \postcode{43017-6221}
}
\email{Jingrui.he@asu.edu}

\author{Jiejun Xu}
% \authornote{The secretary disavows any knowledge of this author's actions.}
\affiliation{%
  \institution{HRL Laboratories, LLC}
%   \streetaddress{P.O. Box 1212}
%   \city{Tempe}
%   \state{Arizona}
%   \postcode{43017-6221}
}
\email{jxu@hrl.com}

%
% By default, the full list of authors will be used in the page headers. Often, this list is too long, and will overlap
% other information printed in the page headers. This command allows the author to define a more concise list
% of authors' names for this purpose.
\renewcommand{\shortauthors}{Trovato and Tobin, et al.}

%
% The abstract is a short summary of the work to be presented in the article.
\begin{abstract}
Graph data widely exist in many high-impact applications. Inspired by the success of deep learning in grid-structured data, graph neural network models have been proposed to learn powerful node-level or graph-level representation. However, most of the existing graph neural networks suffer from the following limitations: (1) there is limited analysis regarding the graph convolution properties, such as {\em seed-oriented}, {\em degree-aware} and {\em order-free}; (2) the node's degree-specific graph structure is not explicitly expressed in graph convolution for distinguishing structure-aware node neighborhoods; (3) the theoretical explanation regarding the graph-level pooling schemes is unclear.

To address these problems, we propose a generic degree-specific graph neural network named \demonet\ motivated by Weisfeiler-Lehman graph isomorphism test that recursively identifies 1-hop neighborhood structures. In order to explicitly capture the graph topology integrated with node attributes, we argue that graph convolution should have three properties: {\em seed-oriented}, {\em degree-aware}, {\em order-free}. To this end, we propose multi-task graph convolution where each task represents node representation learning for nodes with a specific degree value, thus leading to preserving the degree-specific graph structure. In particular, we design two multi-task learning methods: degree-specific weight and hashing functions for graph convolution. In addition, we propose a novel graph-level pooling/readout scheme for learning graph representation provably lying in a degree-specific Hilbert kernel space. The experimental results on several node and graph classification benchmark data sets demonstrate the effectiveness and efficiency of our proposed \demonet\ over state-of-the-art graph neural network models.
\end{abstract}

%
% The code below is generated by the tool at http://dl.acm.org/ccs.cfm.
% Please copy and paste the code instead of the example below.
%
% \begin{CCSXML}
% <ccs2012>
%  <concept>
%   <concept_id>10010520.10010553.10010562</concept_id>
%   <concept_desc>Computer systems organization~Embedded systems</concept_desc>
%   <concept_significance>500</concept_significance>
%  </concept>
%  <concept>
%   <concept_id>10010520.10010575.10010755</concept_id>
%   <concept_desc>Computer systems organization~Redundancy</concept_desc>
%   <concept_significance>300</concept_significance>
%  </concept>
%  <concept>
%   <concept_id>10010520.10010553.10010554</concept_id>
%   <concept_desc>Computer systems organization~Robotics</concept_desc>
%   <concept_significance>100</concept_significance>
%  </concept>
%  <concept>
%   <concept_id>10003033.10003083.10003095</concept_id>
%   <concept_desc>Networks~Network reliability</concept_desc>
%   <concept_significance>100</concept_significance>
%  </concept>
% </ccs2012>
% \end{CCSXML}

% \ccsdesc[500]{Computer systems organization~Embedded systems}
% \ccsdesc[300]{Computer systems organization~Redundancy}
% \ccsdesc{Computer systems organization~Robotics}
% \ccsdesc[100]{Networks~Network reliability}

%
% Keywords. The author(s) should pick words that accurately describe the work being
% presented. Separate the keywords with commas.

%
% A "teaser" image appears between the author and affiliation information and the body 
% of the document, and typically spans the page. 
% \begin{teaserfigure}
%   \includegraphics[width=\textwidth]{sampleteaser}
%   \caption{Seattle Mariners at Spring Training, 2010.}
%   \Description{Enjoying the baseball game from the third-base seats. Ichiro Suzuki preparing to bat.}
%   \label{fig:teaser}
% \end{teaserfigure}

%
% This command processes the author and affiliation and title information and builds
% the first part of the formatted document.
\maketitle

\input{samplebody-conf}

\bibliographystyle{ACM-Reference-Format}
\bibliography{sample-bibliography}
\clearpage
\input{008appendix.tex}

\end{document}

%% file: samplebody-conf.tex
\section{Introduction}
Nowadays, graph data is being generated across multiple high-impact application domains, ranging from bioinformatics~\cite{gilmer2017neural} to financial fraud detection~\cite{zhou2018sparc, zhou2017local}, from genome-wide association study~\cite{wu2018leveraging} to social network analysis~\cite{hamilton2017inductive}. In order to leverage the rich information in graph-structured data, it is of great importance to learn effective node or graph representation from both node/edge attributes and the graph topological structure. To this end, numerous graph neural network models have been proposed recently inspired by the success of deep learning architectures on grid-structured data (e.g., images, videos, languages, etc.). One intuition behind this line of approaches is that the topological structure as well as node attributes could be integrated by recursively aggregating and compressing the continuous feature vectors from local neighborhoods in an end-to-end training architecture.

One key component of graph neural networks~\cite{hamilton2017representation, gilmer2017neural} is the graph convolution (or feature aggregation function) that aggregates and transforms the feature vectors from a node's local neighborhood. By integrating the node attributes with the graph structure information using Laplacian smoothing~\cite{li2018deeper, kipf2016semi} or advanced attention mechanism~\cite{velickovic2017graph}, graph neural networks learn the node representation in a low-dimensional feature space where nearby nodes in the graph would share a similar representation. Moreover, in order to learn the representation for the entire graph, researchers have proposed the graph-level pooling schemes~\cite{atwood2016diffusion} that compress the nodes' representation into a global feature vector. The node or graph representation learned by graph neural networks has achieved state-of-the-art performance in many downstream graph mining tasks, such as node classification~\cite{zhang2018graph}, graph classification~\cite{xu2018powerful}, etc.

\begin{figure}[t]
\includegraphics[width = 3.35in]{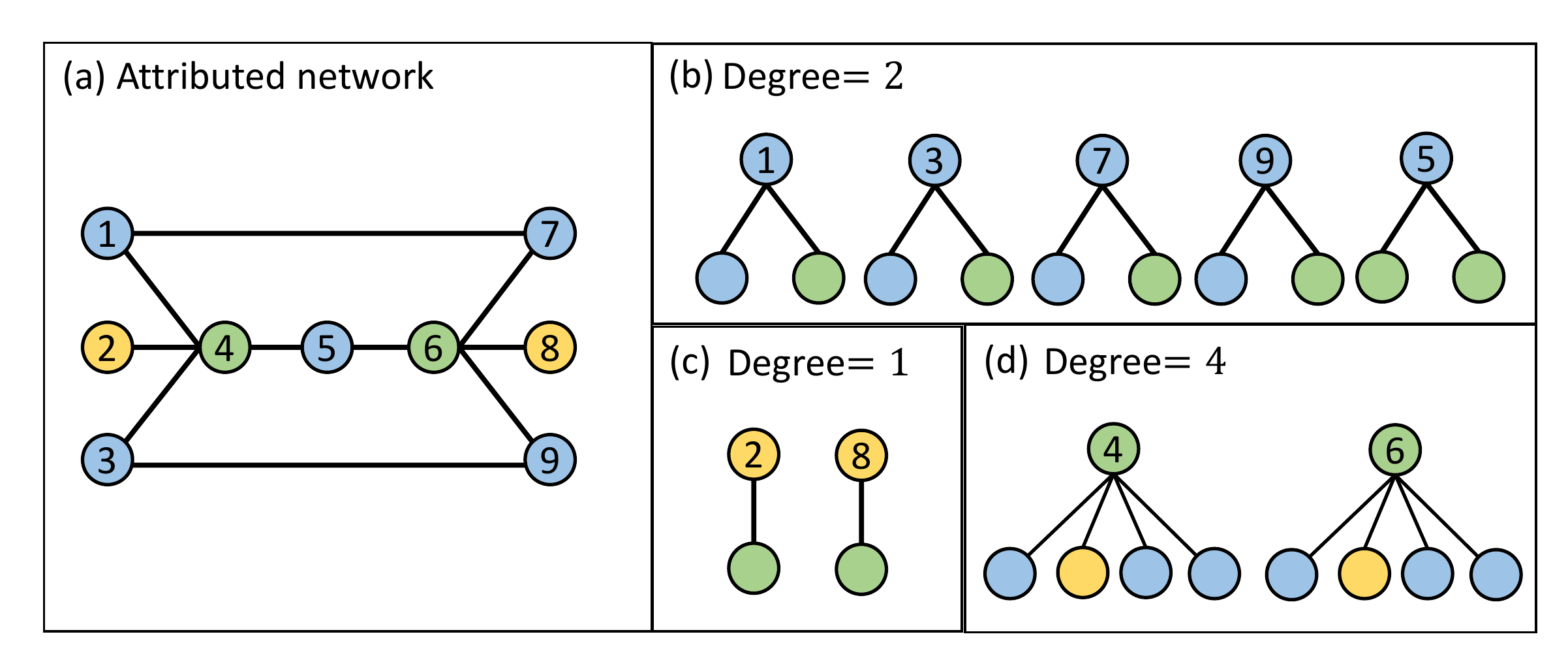}
\caption{Nodes with the same degree value are structurally similar. For example, nodes 1, 3, 5, 7 and 9 in (b), nodes 2 and 8 in (c), nodes 4 and 6 in (d) share similar 1-hop neighborhood structure. Using the proposed model, the learned node representation integrates the degree-specific graph structure and node attributes such that the structurally similar nodes have similar representation.}
\label{degree_struc}
\vspace{-3mm}
\end{figure}

However, most of the existing graph neural networks suffer from the following limitations.
\textbf{ (L1)} There is limited analysis on graph convolution properties that could guide the design of graph neural networks when learning node representation. \textbf{ (L2)} In order to preserve the node proximity, the graph convolution applies a special form of Laplacian smoothing~\cite{li2018deeper}, which simply mixes the attributes from node's neighborhood. This leads to the loss of degree-specific graph structure information for the learned representation. An illustrative example is shown in Figure \ref{degree_struc}: although nodes 4 and 5 are structurally different, they would be mapped to similar representation due to first-order node proximity using existing methods. Moreover, the neighborhood sub-sampling methods used to improve model efficiency~\cite{hamilton2017inductive} significantly degraded the discrimination of degree-specific graph structure. \textbf{ (L3)} The theoretical explanation regarding the graph-level pooling schemes is largely missing.

To address the above problems, in this paper, we propose a generic graph neural network model \demonet\ that considers the degree-specific graph structure in learning both node and graph representation. Inspired by Weisfeiler-Lehman graph isomorphism test \cite{weisfeiler1968reduction}, the graph convolution of graph neural networks should have three properties: {\em seed-oriented}, {\em degree-aware}, {\em order-free}, in order to map different neighborhoods to different feature representation. 
As shown in Figure \ref{degree_struc}, nodes with identical degree value typically share similar subtree (root node followed by its 1-hop neighbors) structures. As a result, the representation of nodes 2 and 8 should be close in the feature space due to the similar subtree structure. On the other hand, nodes 4 and 5 have different subtree structures (i.e., number of subtree leaves), and they indicate different roles in the network, e.g., leader vs. deputy in a covert group. Therefore, they should not be mapped closely in the feature space.

To the best of our knowledge, very little effort on graph neural networks is devoted to learning the degree-specific representation for each node or the entire graph. To bridge the gap, we present a degree-specific graph convolution by assuming that nodes with the same degree value would share the same graph convolution. It can be formulated as a multi-task feature learning problem where each task represents the node representation learning for nodes with specific degree values. 

In addition, we introduce a degree-specific graph-level pooling scheme to learn the graph representation. We theoretically show that the graph representation learned by our model lies in a Reproducing Kernel Hilbert space (RKHS) induced by a degree-specific Weisfeiler-Lehman graph kernel. The most similar work to us is Graph Isomorphism Network (GIN) \cite{xu2018powerful} which used the sum-aggregator associated with multi-layer perceptrons as the neighborhood-injective graph convolution that mapped different node neighborhood to different features. However, one issue of GIN is that 
the degree-aware structures are implicitly expressed in its graph convolution relying on the universal approximation capacity of multi-layer perceptrons.

The main contributions of this paper are summarized as follows:
\begin{enumerate}
    \item We provide theoretical analysis for graph neural networks from the perspective of Weisfeiler-Lehman graph isomorphism test, which motivates us to design the graph convolution based on the following properties: seed-oriented, degree-aware and order-free.
    
    \item we propose a generic graph neural network framework named \demonet\ by assuming that nodes with the same degree value would share the same graph convolution. A degree-specific multi-task graph convolution function is presented to learn the node representation. Furthermore, a novel graph-level pooling scheme is introduced for learning the graph representation provably lying in a degree-specific Hilbert kernel space. 
    
    \item The experimental results on several node and graph classification benchmark data sets demonstrate the effectiveness and efficiency of our proposed \demonet\ model.
\end{enumerate}

The rest of the paper is organized as follows. We review the related work in Section 2, followed by the problem definition and background introduction in Section 3. Section 4 presents our proposed \demonet\ framework for node and graph representation learning. The extensive experiments and discussion are provided in Section 5. Finally, we conclude the paper in Section 6.

\section{Related Work}
In this section, we briefly review the related work on graph neural networks for node and graph classification.

\subsection{Node Classification}
Most of the existing graph neural networks \cite{hamilton2017representation} learn the node representation by recursively aggregating the continuous feature vectors from local neighborhoods in an end-to-end fashion. They could be fitted into the Message Passing Neural Networks (MPNNs) \cite{gilmer2017neural} which explained the feature aggregation of graph neural networks as message passing in local neighborhoods.
Generally, they focus on extracting the spatial topological information by operating the convolutions in the node domain \cite{zhang2018graph}, which differs from some spectral approaches \cite{defferrard2016convolutional, kipf2016semi} considering a node representation in the spectral domain. Graph Convolutional Network (GCN) \cite{kipf2016semi} defined the convolution operation via a neighborhood aggregation function. Following the same intuition, many graph neural network models have been proposed with different aggregation functions, e.g., attention mechanism \cite{velickovic2017graph}, mean and max functions \cite{hamilton2017inductive}, etc.

However, most of the graph neural network architectures are motivated by the success of deep learning on grid-like data, thus leading to little theoretical analysis for explaining the high performance and guiding the novel methodologies. Up till now, some work have been proposed to explain why graph neural networks work. The convolution of GCN was a special form of Laplacian smoothing on graph \cite{li2018deeper}, which explained the over-smoothing phenomena brought by many convolution layers. Lei et al. \cite{lei2017deriving} showed that the graph representation generated by graph neural networks lies in the Reproducing Kernel Hilbert Space (RKHS) of some popular graph kernels. Moreover, it shows that 1-dimensional aggregation-based graph neural networks are at most as powerful as the Weisfeiler-Lehman (WL) isomorphism test \cite{weisfeiler1968reduction} in distinguishing graphs \cite{xu2018powerful}. Compared with the existing work on graph neural networks, in this paper, we design a degree-specific graph convolution that captures the node neighborhood structures inspired by WL isomorphism test. This is in sharp contrast to the existing work which focused on preserving the node proximity in the feature space, thus leading to the loss of local graph structures.

\subsection{Graph Classification}
The graph-level pooling/readout schemes aim to learn a representation of the entire graph from its node representations for graph-level classification tasks. Mean/max/sum functions are commonly used due to its computational efficiency and effectiveness \cite{atwood2016diffusion, xu2018powerful}. One challenge for graph-level pooling is to maintain the invariance to node order. PATCHY-SAN \cite{niepert2016learning} first adopted the external software to obtain a global node order for the entire graph, which is very time-consuming. More recently, a number of graph neural network models have been proposed \cite{zhang2018end, xu2018powerful, ying2018hierarchical}, which formulated the node representation learning and graph-level pooling into a unified framework. Different from graph kernel approaches \cite{shervashidze2011weisfeiler, yanardag2015deep} that intuitively extract the graph feature or define the graph similarity using ad-hoc knowledge or random walk properties, graph neural networks would automatically learn the graph representation to integrate node attributes with its topological information via an end-to-end training architecture.

Nevertheless, very little effort has been devoted to explicitly considering the degree-specific graph structures for graph representation learning. Our proposed degree-specific graph-level pooling method is designed to address this issue by compressing the learned node representation according to degree values.

\section{Preliminaries}
In this section, we introduce the notation and problem definition, as well as some background information on graph neural networks.

\subsection{Notation}
Suppose that a graph is represented as $G = (V, E)$, where $V=\{v_1, \cdots, v_n\}$ is the set of $n$ nodes and $E \subseteq V \times V$ is the edge set. Let $X \in \mathbb{R}^{n \times D}$ denote the attribute matrix where each row $x_v$ is the $D$-dimensional attribute vector for node $v$. The graph $G$ can also be represented by an adjacency matrix $A\in \mathbb{R}^{n\times n}$, where $A_{ij}$ represents the similarity between $v_i$ and $v_j$ on the graph. For each node $v \in V$, its 1-hop neighborhood is denoted as $N(v)$.
Let $\mathcal{G} = \{G_1, \cdots, G_t\}$ denote a set of graphs. In this paper, we focus on undirected attributed networks, although our model can be naturally generalized to other types of networks. The main notation used in this paper is summarized in Table \ref{tab: notations}.

\begin{table}[t]
\centering
\small
\caption{Notation} \label{tab: notations}
\begin{tabular}{|c|l|}
\hline
\multicolumn{1}{|c|}{Notation} & \multicolumn{1}{c|}{Definition}         \\ \hline
$\mathcal{G} = \{G_i\}_{i=1}^t$ &  A set of graphs \\ \hline
$G=(V, E)$ & A graph $G$ with node set $V$ and edge set $E$ \\ \hline
$X$ & Attribute matrix \\ \hline
$A$ & Adjacency matrix \\ \hline
$n$ & Number of nodes in the graph \\ \hline
$d$ & Dimensionality of the node or graph representation \\ \hline
$N(v)$ & 1-hop neighborhood of node $v$ \\ \hline
$\mathcal{I}_{G}$ & Indices of labeled nodes' for node classification \\ \hline
$\mathcal{I}_{\mathcal{G}}$ & Indices of labeled graphs' for graph classification \\ \hline
$y_v$ & Label of node $v$ \\ \hline
$\hat{y}_i$ & Label of graph $G_i$ \\ \hline
$h_v^k$ & Node $v$'s representation at the $k^{\textrm{th}}$ iteration\\ \hline
$h_{N(v)}$ & Feature set within node $v$'s neighborhood \\  \hline
$\mathcal{T}$ & A set of subtrees\\ \hline
$degree(G)$ &  A set of the degree values in graph $G$ \\ \hline
\end{tabular}
\vspace{-3mm}
\end{table}

\subsection{Problem Definition}
In this paper, we focus on two problems: node-level and graph-level representation learning by formulating a novel degree-specific graph neural network model. Furthermore, we analyze the proposed model from various aspects, and empirically demonstrate its superior performance on both node and graph classification.

Formally, the node- and graph-level representation learning problems can be defined below.
\begin{definition}(\textit{\textbf{Node-level Representation Learning}}) \\
\indent \textbf{Input:} (i) An attributed graph $G=(V,E)$ with adjacency matrix $A \in \mathbb{R}^{n\times n}$ and node attributes $X\in \mathbb{R}^{n \times D}$; (ii) Labeled training nodes $\{x_v, y_v\}_{v \in \mathcal{I}_{G}}$. \\
\indent \textbf{Output:} A vector representation $e_v \in \mathbb{R}^{d}$ for each node $v\in V$ on the $d$-dimensional embedding space where nodes would be well separated if their local neighborhoods are structurally different.
\end{definition}

\begin{definition}(\textit{\textbf{Graph-level Representation Learning}}) \\
\indent \textbf{Input:} (i) A set of attributed graphs $\mathcal{G}= \{G_i\}_{i=1}^t$ with adjacency matrix $A_i \in \mathbb{R}^{n_i\times n_i}$ and node attributes $X_i \in \mathbb{R}^{n_i \times D}$; (ii) Labeled training graphs $\{G_i, \hat{y}_i\}_{i \in \mathcal{I}_{\mathcal{G}}}$. \\
\indent \textbf{Output:} A vector representation $g_i \in \mathbb{R}^{d}$ for each graph $G_i$ on the $d$-dimensional embedding space where graphs would be well separated if they have different graph topological structure.
\end{definition}

\subsection{Graph Neural Networks}
\begin{figure}
\centering
\includegraphics[width = 3.35in]{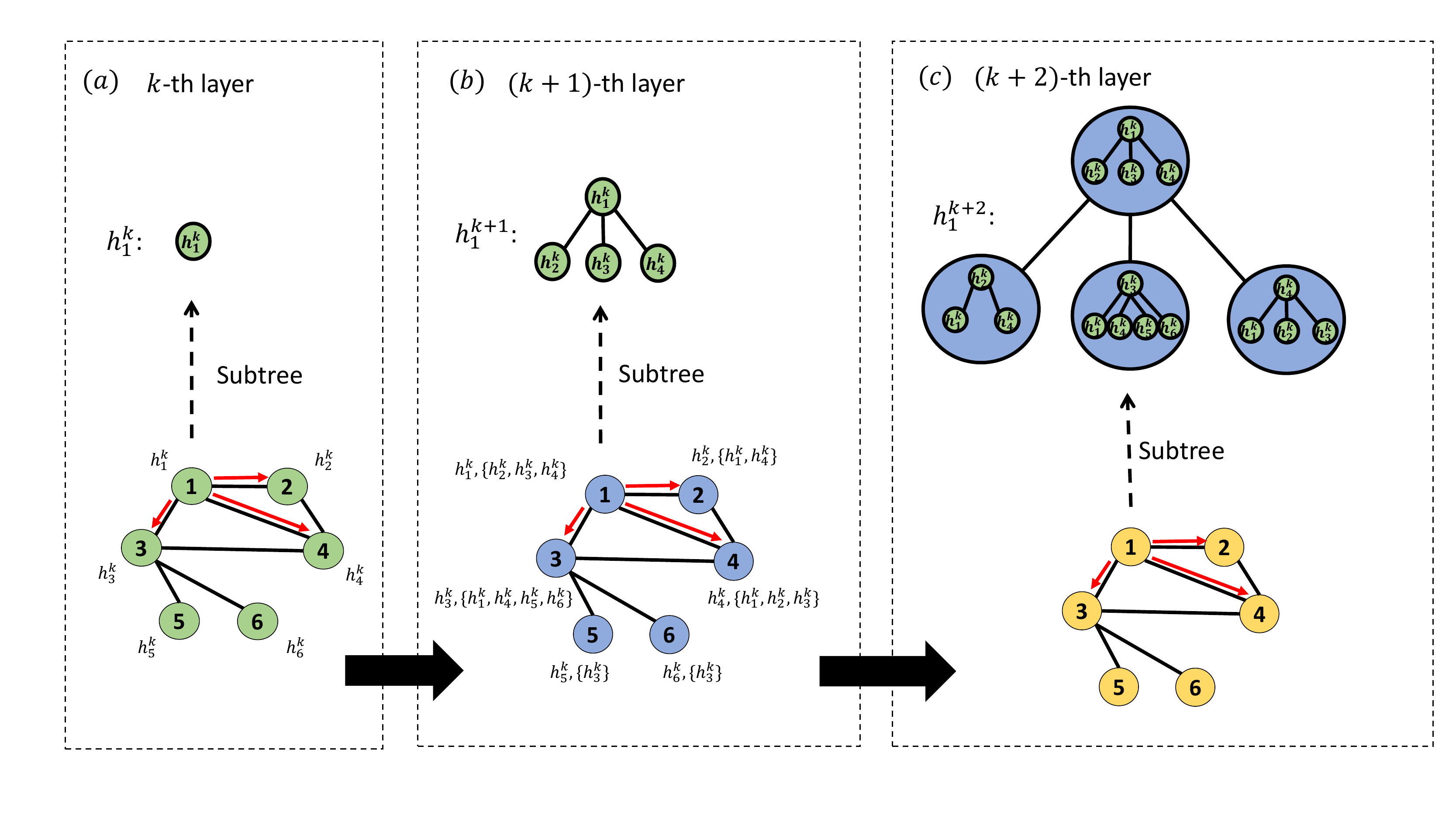}
\caption{Feature aggregation of the graph neural networks: For node 1, its feature is (a) $h_1^{k}$ at $k^{\textrm{th}}$ layer; (b) $h_1^{k+1}$ at $(k+1)^{\textrm{th}}$ layer compressed from a subtree $(h_1^k; \{h_2^k, h_3^k, h_4^k\})$; (c) $h_1^{k+2}$ at $(k+2)^{\textrm{th}}$ layer learned from a subtree $(h_1^{k+1}; \{h_2^{k+1}, h_3^{k+1}, h_4^{k+1}\})$.}
\label{subtree}
\vspace{-3mm}
\end{figure}

It has been observed that a broad class of graph neural network (GNN) architectures followed the 1-dimensional Weisfeiler-Lehman (WL) graph isomorphism test \cite{weisfeiler1968reduction}. From the perspective of WL isomorphism test, they mainly consist of the following crucial steps at each iteration of feature aggregation:
\begin{itemize}
\item Feature initialization (label\footnote{ Here, label is an identifier of nodes. In order not to be confused with a class label, we will use node attribute to represent it in this paper.} initialization): The node features are initialized by original attribute vectors.
\item Neighborhood detection (multiset-label determination): It decides the local neighborhood in which node gathers the information from neighbors. More specifically, a seed\footnote{The seed denotes the root node to be learned in the graph. For example, node $v_1$ in Figure \ref{subtree} is a seed when updating its feature at each iteration.} followed by its neighbors generates a subtree pattern.
\item Neighbors sorting (multiset-label sorting): The neighbors are sorted in the ascending or descending order of degree values. The subtrees with permutation order of neighbors are recognized as the same one.
\item Feature aggregation (label compression): The node feature is updated by compressing the feature vectors of the aggregated neighbors including itself.
\item Graph-level pooling (graph representation): It summarizes all the node features to form a global graph representation.
\end{itemize}

Next, we briefly go over some existing graph neural network models, which follow the aforementioned steps of the 1-dimensional WL algorithms. We would like to point out that graph neural networks would learn the node or graph representation using continuous node attributes, whereas WL algorithms update the node attributes by directly compressing the augmented discrete attributes.

Taking 1-hop neighborhood $N(v) = \{u | (v, u) \in E\}$ into consideration at each iteration, the following node-level graph neural network variants have the same feature initialization and neighborhood detection on learning node representation. And when element-wise average or max operations are used for feature aggregation, graph neural networks would be invariant to the order of neighbors. We summarize the feature aggregation functions (graph convolution) of those graph neural networks as follows.

\begin{itemize}
\item Graph Convolutional Network (GCN)~\cite{kipf2016semi}:
\begin{equation}
h_v^{k} = \sigma \Big(\sum\nolimits_{u \in \{v\} \cup N(v)} \widehat{a}_{vu} W^{k} h_u^{k-1} \Big)
\end{equation}
where $\widehat{A} = (\widehat{a}_{vu}) \in \mathbb{R}^{n\times n}$ is the re-normalization of the adjacency matrix $A$ with added self-loops, and $W^{k}$ is the trainable matrix at $k^{\textrm{th}}$ layer. It is essentially a weighted feature aggregation from node neighborhood.
\item Graph Attention Network (GAT)~\cite{velickovic2017graph}:
\begin{equation}
h_v^{k} = \sigma \Big(\sum\nolimits_{u \in \{v\} \cup N(v)} \alpha_{vu} W^k h_u^{k-1} \Big)
\end{equation}
where $\alpha_{vu}$ is a self-attention score indicating the importance of node $u$ to node $v$ on feature aggregation. It is obvious that GCN can be considered as a special case of GAT when the attention score $\alpha_{vu}$ is defined as $\widehat{a}_{vu}$.
\item GraphSAGE~\cite{hamilton2017inductive}:
\begin{equation}
\begin{split}
h_{N(v)}^k &= AGGREGATE_k \Big(\{h_u^{k-1}|u \in N(v)\} \Big) \\
h_v^k &= \sigma \Big(W^k \cdot CONCAT(h_v^{k-1}, h_{N(v)}^k) \Big)
\end{split}
\end{equation}
where mean-, max- and LSTM-aggregator are presented for feature aggregation. Though LSTM considers node neighbors as an ordered sequence, the LSTM aggregator is adapted on an unordered neighbors with random permutation.
\end{itemize}

There are some observations from these GNN variants: (i) Their feature aggregation schemes are invariant to the order of the neighbors except for GraphSAGE with LSTM-aggregator; (ii) The output feature at $k$-layer neural network can be seen as the representation of a subtree around the seed; (iii) The node representation become closer and indistinguishable when the neural layers are going deeper, because the subtrees would share more common elements. However, little work theoretically discusses the reasons behind these observations to guide the design of graph neural networks: how is the node representation affected by node degree and order of neighbors? what kind of graph convolution is required to learn the subtree structures? Inspired by WL graph isomorphism test, we present a degree-specific graph neural network model named \demonet\ in Section 4 to discuss those problems.

Additionally, the neighborhood aggregation schemes of graph neural networks, such as mean-aggregator in GraphSAGE~\cite{hamilton2017inductive}, self-attention in GAT~\cite{velickovic2017graph}, can be regarded as the relabeling step in WL isomorphism test. Figure \ref{subtree} provides an example to illustrate the essence of feature aggregation on graph neural networks. The node feature is actually a special representation of subtree consisting of the seed followed by its neighbors. For example, node 1's feature $h_1^{k+1}$ represents the subtree $(h_1^k; \{h_2^k, h_3^k, h_4^k\})$ collected from previous layer. As a result, graph neural networks with $k$ layers learn the representation of subtree with depth $k$ rooted at the seed. That provides us an intuition to design a graph convolution for explicitly preserving the degree-specific subtree structures.

\section{Proposed Model: \demonet}

In this section, we propose a generic degree-specific graph neural network named \demonet. Key to our algorithm is the degree-specific graph convolution for feature aggregation which can map different subtrees to different feature vectors. Figure \ref{framework} provides an overview of the proposed \demonet\ framework on learning node and graph representation, which will be described in detail below.

\subsection{Node Representation Learning}
Let $h_{N(v)}$ denote the feature set $\{h_u | u \in N(v)\}$ within node $v$'s neighborhood.
Let $\mathcal{T} = \big\{ \big(h_v, h_{N(v)} \big) \big\}$ be the set of subtrees consisting of the features of seed $v$ and its 1-hop neighbors $N(v)$. To formalize our analysis, we first give the definition of structurally identical subtree below.
% \he{This definition should be moved to before all the discussions about structurally identical subtrees.}
\begin{definition}(Structurally Identical Subtree)
Any two subtrees in $\mathcal{T}$ are structurally identical if the only possible difference between them is the order of neighbors.
\end{definition}

The following lemma shows that graph neural networks could distinguish the local graph structures as well as the WL graph isomorphism test when \textbf{graph convolution is an injective function} that maps two subtrees in $\mathcal{T}$ to different features if they are not structurally identical.
% \he{Can you summarize what the following lemma is trying to prove here?}

\begin{lemma}
Let $G=\{V, E\}$ be a graph and $u, v \in V$ be two nodes in the graph. When the mapping function $f: \mathcal{T} \rightarrow \mathbb{R}^d$ in graph neural networks is injective, the learned features of $v$ and $u$ will be different if and only if the WL graph isomorphism test determines that they are not structurally identical.
\end{lemma}
% \begin{proof}
% If the Weisfeiler-Lehman graph isomorphism test determines that $u$ and $v$ are not structurally identical, there exist the number of iterations $k$ such that the subtree structure around the seeds $u$ and $v$ are different after $k$ iterations. Then the injective mapping function $f$ will map them to different embedding representations.

% If the function $f$ maps $u$ and $v$ to different embedding at $k$-th iteration, the current subtree structures around $u$ and $v$ are not identical. That is, their $k$-hop neighborhoods are not structurally identical.
% \end{proof}

The feature aggregation of graph neural networks can be simply summarized as follows.
\begin{equation}
    h_v^k = f(\{ h_v^{k-1}, h_u^{k-1} | u \in N(v) \})
    \label{G1}
\end{equation}
Obviously, most of the existing graph neural networks \cite{kipf2016semi, velickovic2017graph} did not consider the injective aggregation function when learning node representation. From the perspective of WL isomorphism test, an injective graph convolution has the following properties.

\begin{lemma} (Properties)
Let $f: \mathcal{T} \rightarrow \mathbb{R}^d$ be the aggregation function. If it is an injective function that maps any different subtrees in $\mathcal{T}$ to different feature vectors, then it has the following properties:
\begin{enumerate}[label=(\roman*)]
\item Seed-oriented: $f\big(h_i, \{h_u | u \in N(i)\} \big) \ne f\big(h_j, \{h_w | w \in N(j)\} \big)$ if the seeds' attributes are different, i.e., $h_i \ne h_j$.
\item Degree-aware: $f\big(h_i, \{h_u | u \in N(i)\} \big) \ne f\big(h_j, \{h_w | w \in N(j)\} \big)$ if the seeds' degree values are different, i.e., $deg(i) \ne deg(j)$.
\item Order-free: $f\big(h_i, \{h_u | u \in N(i)\} \big) = f\big(h_j, \{h_w | w \in N(j)\} \big)$ if $h_i = h_j$ and the only possible difference between $\{h_u | u \in N(i)\}$ and $\{h_w | w \in N(j)\}$ is the order of neighbors.
\end{enumerate}
\end{lemma}
% \begin{proof}
% If the function $f$ is injective, it is easy to show that when two subtrees have different seed features or degrees, they are structurally different. Thus they would be mapped to different feature vectors as described in (i) and (ii). And if the only difference of two subtrees is the order of neighbors, they essentially represent the same local structure in the graph, thus leading to be mapped to the same feature vector shown in (iii).
% \end{proof}

\begin{figure}
\includegraphics[width = 1.5in]{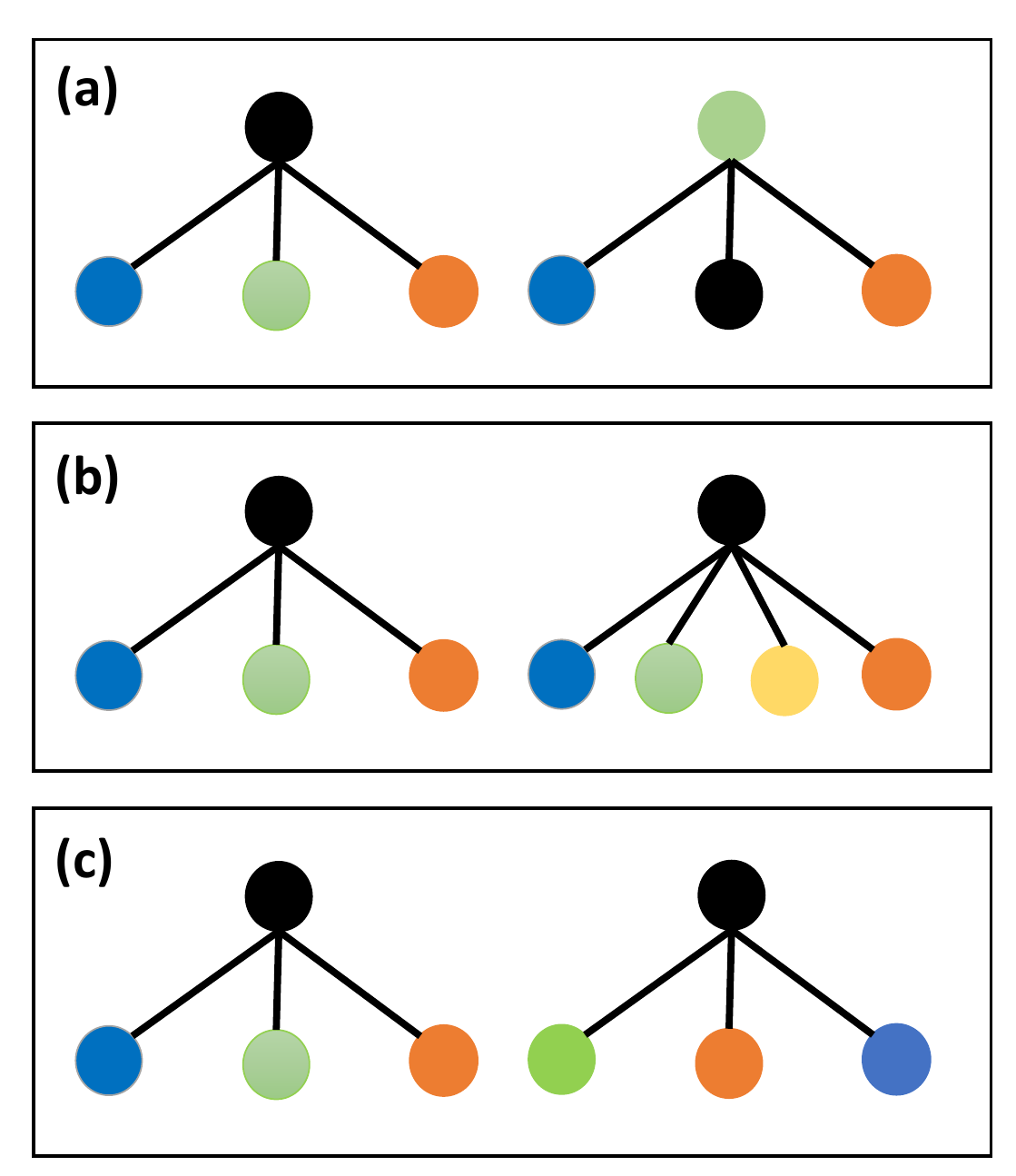}
\caption{Examples of subtree in $\mathcal{T}$ with: (a) different seeds' attributes; (b) different seeds' degree values; (c) different neighbors' order. In such cases, two subtrees in (a) and (b) are mapped to different feature vectors, respectively. Two subtrees in (c) will be mapped to the same feature vector. (Best seen in color. Colors denote the node attributes.)}
\label{mapping}
\vspace{-3mm}
\end{figure}

Figure \ref{mapping} lists some examples to illustrate those properties. The injective function $f$ maps the subtrees in Figure \ref{mapping}(a) to different features due to the distinctive seeds' attributes. Here, we hold that the subtree's structure properties are guided by seed node. Thus they are not structurally identical though both subtrees share the same leaf elements. Seeds' degree values also decide the subtree structure (shown in Figure \ref{mapping}(b)) because it is obvious that nodes with identical degree value share the similar structure. Figure \ref{mapping}(c) shows that neighbors' order will not change the subtree structure. 

These properties will guide us to build a structure-specific graph neural network model.
Based on properties (i) and (ii), the feature aggregation function in Eq. (\ref{G1}) can be expressed as follow.
\begin{equation}
    h_v^k = f_s(h_v^{k-1}) \circ f_{deg(v)}\big( \{h_u^{k-1} | u \in N(v) \} \big)
    \label{G2}
\end{equation}
where $f_s$ and $f_{deg(v)}$ are seed-related and degree-specific mapping functions, respectively, and $deg(v)$ denotes the degree value of node $v$. All the nodes share one seed-oriented mapping function $f_s$, but have a degree-specific function for compressing node neighborhoods. Here, $\circ$ denotes the vector concatenation which combines the mapped features to form a single vector. If $f_s$ and $f_{deg(v)}$ are injective, it will have the first two properties in Lemma 4.2 that subtrees with different seeds' features or degree values would be mapped differently. Additionally, the degree-specific mapping function $f_{deg(v)}$ should be symmetric\footnote{A symmetric function of $n$ variables is one whose value given $n$ arguments is the same no matter the order of the arguments. For example, $f(x_1, x_2) = f(x_2, x_1)$ for any pair $(x_1, x_2)$.} that is invariant to the order of neighbors. And we have the following theorem (proven in Appendix) to show the existence of mapping functions $f_s$ and $f_{deg}$.
\begin{theorem}(Existence Theorem)
\label{T: existence}
Assume $\mathcal{T}$ is countable, there exist mapping functions $f_s$ and $\{ f_{deg} | deg \in degree(G) \}$ such that for any two subtrees in $\mathcal{T}$, the function $f: \mathcal{T} \rightarrow \mathbb{R}^d$ defined in Eq. (5) maps them to different features if they are not structurally identical.
\end{theorem}

Next, we present our graph neural network model where the injective aggregation function could be approximated by multi-layer neural network due to its exceptional expression power. For seed-related mapping function $f_s(\cdot)$ in Eq. (5), we use a simple one-layer fully-connected neural network as follows.
\begin{equation}
    f_s(h_v^{k-1}) = \sigma(W_0^k h_v^{k-1})
\end{equation}
where the trainable matrix $W_0^k$ is shared by all the seeds at $k^{\textrm{th}}$ hidden layer. Here $\sigma(\cdot)$ is a nonlinear activation function. 

\begin{figure*}
\centering
\includegraphics[width = 5in]{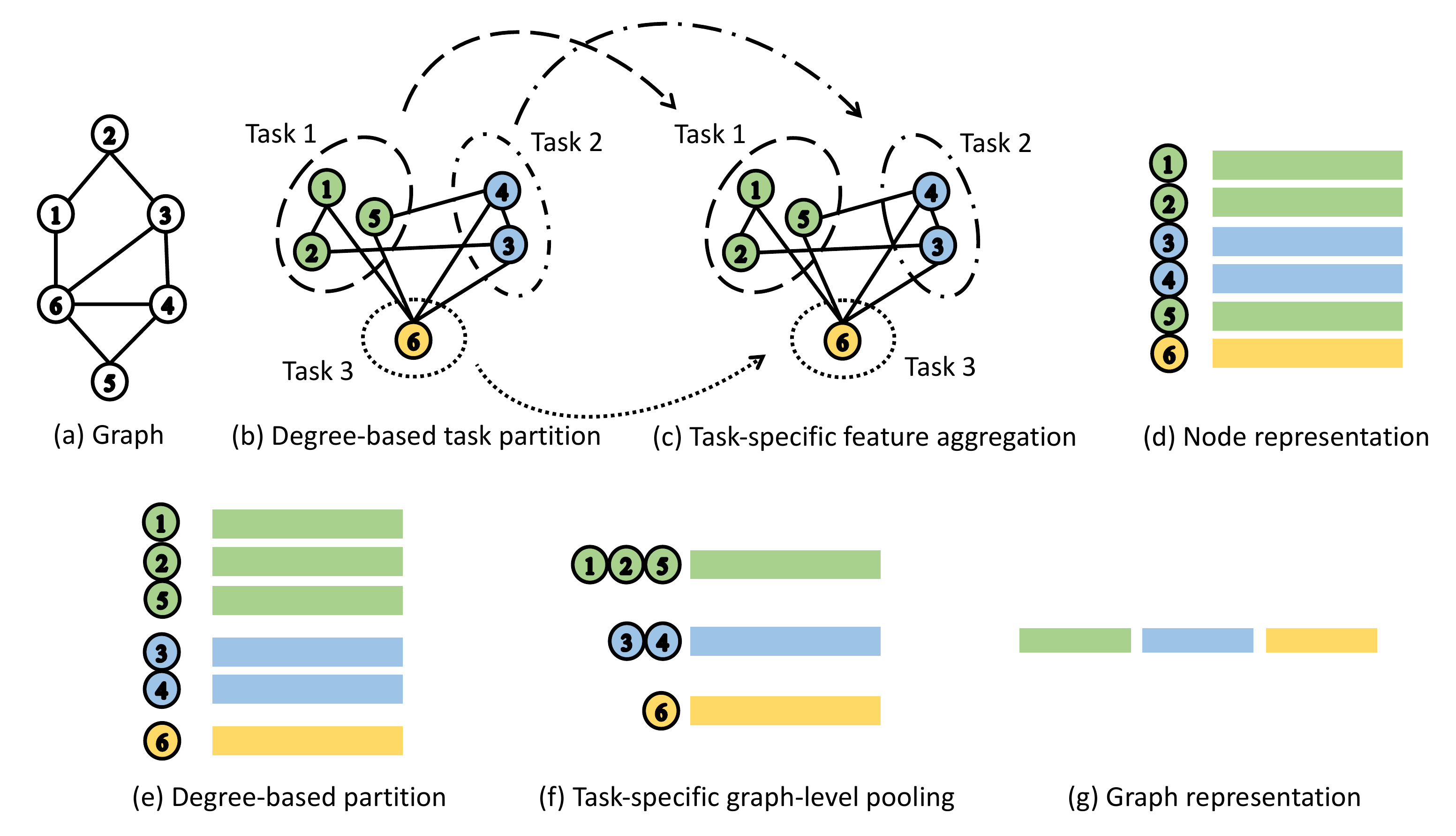}
\caption{Overview of our proposed \demonet\ framework (best seen in color). (b) and (c) represent the multi-task feature aggregation. The node representation in (d) can be used for node-level classification. For learning graph embedding, (e)-(g) provide the graph-level pooling method based on node degree distribution for learning graph representation.}
\label{framework}
\vspace{-3mm}
\end{figure*}

For degree-specific neighborhood aggregation on $h_{N(v)}$, it can be formulated as a multi-task feature learning problem (shown in Figure \ref{framework}(b)(c)) in which each task represents node representation learning for nodes with a specific degree value, thus leading to preserving the degree-specific graph structure. Here, we present two schemes for this multi-task learning problem.

\noindent\textbf{Degree-specific weight function: } The degree-specific aggregation function can be expressed as follow.
\begin{equation}
\begin{aligned}
    f_{deg(v)}(h_{N(v)}^{k-1})&=\sigma \Big(\sum\nolimits_{u\in N(v)} (W_g^k + W_{deg(v)}^k) h_u^{k-1} \Big)
\end{aligned}
\end{equation}
where $W_{deg(v)}^k$ is a degree-specific trainable matrix at $k^{\textrm{th}}$ layer and $W_g^k$ is a global trainable matrix shared by all the seeds.

\noindent\textbf{Hashing function:} Since the number of degree values on graphs could be very large, a critical challenge is how to perform multi-task learning efficiently. To address this challenge, hash kernel~\cite{weinberger2009feature} (also called feature hashing or hash trick) is applied for our multi-task neighborhood learning problem. Given two vectors $x$ and $x'$, the hash map $\phi$ and the corresponding kernel $K_{\phi}(\cdot,\cdot)$ are defined:
\begin{equation}
\phi_i^{\xi_1,\xi_2}(x) = \sum\nolimits_{j: \xi_1(j)=i} \xi_2 (j) x_j
\end{equation}
\begin{equation}
K_{\phi}(x, x') = \left\langle {x, x'} \right\rangle _{\phi} = \left\langle \phi^{\xi_1,\xi_2}(x), \phi^{\xi_1,\xi_2}(x') \right\rangle
\end{equation}
where $\xi_1$ and $\xi_2$ denote two hash functions such that $\xi_1: \mathbb{N} \rightarrow \{1, \cdots, m\}$ and $\xi_2 : \mathbb{N} \rightarrow \{1,-1\}$. Notice that hash kernel is unbiased, i.e., $E_{\phi} [ \left\langle{x,x'}\right\rangle_{\phi} ] = \left\langle{x,x'}\right\rangle$ for any pair of input feature vectors. Let $w_{deg(v)}^k$ denote one of the row vectors in $W_{deg(v)}^k$, then we have $E_{\phi} \Big[ \left\langle{w_{deg(v)}^k, h_u^{k-1}}\right\rangle_{\phi} \Big] = \left\langle{w_{deg(v)}^k,  h_u^{k-1}}\right\rangle$. In this way, the multi-task feature aggregation function $f_{deg(v)}$ can be expressed as:
\begin{equation}
    f_{deg(v)}(h_{N(v)}^{k-1}) = \sigma \Big(\sum_{u\in N(v)} W^k \big( \phi_g(h_u^{k-1}) + \phi_{deg}(h_u^{k-1}) \big) \Big)
\end{equation}
where $W^k = \phi_g(W_g^k) + \sum_{deg} \phi_{deg}(W_{deg}^k)$ is the trainable matrix shared by all the nodes, and $\phi_g(\cdot)$ and $\phi_{deg}(\cdot)$ are global and degree-specific hash maps, respectively. 

One common assumption in multi-task learning is that all the tasks are related with some shared knowledge, and meanwhile have their own task-specific knowledge. As shown in Figure \ref{mapping}(b), two subtrees are structurally different, but they share some common leaves for neighborhood aggregation. By adopting both common (global) and task-specific (local) weight/hash functions, it allows learning the shared sub-structures and degree-specific neighborhood structures simultaneously.

There might be many different node degrees in real networks. One intuitive idea is that we could partition the degree values into several buckets to reduce the number of tasks. This heuristic solution might improve our model robustness to noisy graph structure or labeled nodes on source networks brought by human annotations \cite{zhou2016crowdsourcing, zhou2017multic2}.
We leave this as our future work because hashing kernel \cite{weinberger2009feature} used in \demonet\ is efficient to tackle large-scale multi-task learning problem.

\subsection{Graph Representation Learning}
The goal of graph representation learning is to use a compact feature vector to represent the entire graph. To this end, we provide a degree-specific graph-level pooling scheme.

When graph neural networks are going deeper, node representation actually captures the higher-order topological information within its local neighborhood. By mapping the original graph to a sequence of graphs $\{G_0, G_1, \cdots, G_K\}$ where $G_0$ denotes the original graph and $G_k (1 \leq k \leq K)$ represents the graph after the $k^{\textrm{th}}$ layer of feature aggregation (as shown in Figure \ref{framework}(e)-(g)), the $k^{\textrm{th}}$ graph representation can be expressed as follow.
\begin{equation}
    h_{G_k} = CONCAT \Big(\big\{\sum\nolimits_{v \in V}h_v^k \cdot \delta(deg(v), d) \big\} \big|d \in degree(G) \Big)
\end{equation}
% \he{What is $degree(G)$?}
where $degree(G)$ denotes the set of degree values in graph $G$, and $\delta(\cdot,\cdot)$ is 1 when its two arguments are equal and 0 otherwise.

As discussed before, the node representation in $G_k$ captures the topological information within $k$-hop neighborhood. In order to consider all the subtrees' information, we concatenate the representation $h_{G_k}$ from all graphs $\{G_0, G_1, \cdots, G_K\}$:
\begin{equation}
    h_G = CONCAT \big(~~~ h_{G_k} ~~~| k=0,1,\cdots,K \big)
\end{equation}

% The intuition behind our degree-specific pooling method is that nodes with identical degree value will be more likely to be matched for graph isomorphism test. Based on Lemma 4.2 and Theorem 4.4, given a sequence of graphs $\{G_0, G_1, \cdots, G_K\}$ above, if two nodes $u$ and $v$ have different degree values, then they will not be structurally identical in any graph of $\{G_0, G_1, \cdots, G_K\}$. The pooling schemes with simply mixing all the node feature vectors \cite{atwood2016diffusion} would together would lead to the loss of degree-specific information. Instead, compressing node feature vectors according to degrees could make the graph-level representation distinctive if they have different degree distribution.

Next, we compare the degree-specific pooling scheme with existing graph-level pooling methods~\cite{atwood2016diffusion, xu2018powerful} and Weisfeiler-Lehman (WL) subtree kernel~\cite{shervashidze2011weisfeiler}.
We define a degree-specific WL kernel:
\begin{equation}
\begin{aligned}
    \mathcal{K}_{DWL}(G_k, G'_k) &= \left\langle{\phi_{DWL}(G_k), \phi_{DWL}(G'_k) }\right\rangle \\
    &= \sum_{v \in V}\sum_{v' \in V'} \delta(deg(v), deg(v')) \cdot \left\langle{h_v^k, h_{v'}^k }\right\rangle 
\end{aligned}
\end{equation}
The corresponding mapping function is defined as:
\begin{equation}
    \phi_{DWL}(G_k) = \big[ c(G_k,d_1) \circ \cdots \circ c(G_k, d_{|degree(G_k)|}) \big]
\end{equation}
where $d_i$ is the degree of $G_k$, and
\begin{equation}
    c(G_k,d_i) = \sum\nolimits_{v\in V} h_v^k \cdot \delta(d_i, deg(v))
\end{equation}

As shown in \cite{lei2017deriving}, the non-linear activation function $\sigma(\cdot)$ has a mapping function $\phi_{\sigma}(\cdot)$ such that $\sigma(\mathbf{w}^T \mathbf{x})= \left\langle{ \phi_{\sigma(\mathbf{x})},  \psi(\mathbf{w})}\right\rangle$ for some mapping $\psi(\mathbf{w})$ constructed from $\mathbf{w}$.
By the following theorem (proven in Appendix), we show that our graph representation lies in a degree-specific Hilbert kernel space. 
\begin{theorem}
\label{T: kernel}
For a degree-specific Weisfeiler-Lehman kernel, the graph representation $h_G$ in Eq. (12) belongs to the Reproducing Kernel Hilbert Space (RKHS) of kernel $\mathcal{K}_{\sigma,DWL}(\cdot, \cdot)$ where
\begin{equation}
    \mathcal{K}_{\sigma,DWL}(G_k, G'_{k}) = \left\langle{\phi_{\sigma}(\phi_{DWL}(G_k)), \phi_{\sigma}(\phi_{DWL}(G'_k)) }\right\rangle
\end{equation}
\end{theorem}

The sum/mean based graph-level pooling approaches make the learned graph representation lie in the kernel as follow.
\begin{equation}
    \mathcal{K}_{MWL}(G_k, G'_k) = \sum\nolimits_{v \in V}\sum\nolimits_{v' \in V'} \left\langle{h_v^k, h_{v'}^k }\right\rangle 
\end{equation}
And WL subtree Kernel \cite{shervashidze2011weisfeiler} can be expressed as:
\begin{equation}
    \mathcal{K}_{WLsubtree}(G_k, G'_k) = \sum\nolimits_{v \in V}\sum\nolimits_{v' \in V'} \delta(h_v^k, h_{v'}^k)
\end{equation}
It is easy to see that: (1) WL subtree kernel cannot be applied to measure the graph similarity when nodes have the continuous attribute vectors. (2) Our graph-level representation lies in a degree-specific kernel space comparing Eq. (13) with (17), thus leading to explicitly preserving the degree-specific graph structure.

\subsection{Discussion}
We compare the proposed \demonet\ with some existing graph neural networks regarding the properties of graph convolution.

Lemma 4.3 shows that an injective aggregation function has three properties: {\em seed-oriented}, {\em degree-aware}, {\em order-free}. We summarize the properties of graph convolution of GCN \cite{kipf2016semi}, GAT \cite{velickovic2017graph}, GraphSAGE \cite{hamilton2017inductive}, and DCNN \cite{atwood2016diffusion} in Table \ref{propeties_comparison}. It can be seen that: (1) The existing graph neural networks do not have all the three properties. More importantly, none of them capture the degree-specific graph structures. (2) For graphSAGE, it is {\em order-free} when using mean or max aggregator. But graphSAGE with LTSM-aggregator is not {\em order-free} because it considers the node neighborhood as an ordered sequence. (3) Our proposed \demonet\ considers all the properties, and the {\em degree-aware} property in particular allows our model to explicitly preserve the neighborhood structures for node and graph representation learning. In addition, the time complexity of graph convolution of \demonet\ is linear with respect to the number of nodes and edges.

\begin{table}
\centering
\caption{Comparison of graph neural networks}\label{propeties_comparison}
\begin{tabular}{|l|c|c|c|}
\hline
Properties & {\em seed-oriented} & {\em degree-aware}  & {\em order-free} \\ \hline
GCN \cite{kipf2016semi} & $\times$ & $\times$ & \checkmark \\ \hline
GAT \cite{velickovic2017graph} & \checkmark & $\times$ & \checkmark \\ \hline
GraphSAGE \cite{hamilton2017inductive} & \checkmark & $\times$ & --- \\ \hline
DCNN \cite{atwood2016diffusion} & $\times$ & $\times$ & \checkmark \\ \hline
\demonet & \checkmark & \checkmark & \checkmark \\ \hline
\end{tabular}
\vspace{-3mm}
\end{table}

\section{Experimental Results}
In this section, we present the experimental results on real networks. In particular, we  focus on answering the following questions: \\
\textbf{Q1:} Is the proposed \demonet\ algorithm effective on node classification compared to the state-of-the-art graph neural networks? \\
\textbf{Q2:} How does the proposed \demonet\ perform on identifying graph structure compared to structure-aware embedding approaches?\\
\textbf{Q3:} How does the proposed \demonet\ with degree-specific graph-level pooling perform on graph classification task? \\
\textbf{Q4:} Is the proposed degree-specific graph convolution of \demonet\ efficient on learning node representation?

\subsection{Experiment Setup}
\textbf{Data Sets:} We use seven node classification data sets, including four social networks and three air-traffic networks. Facebook, Wiki-Vote \cite{snapnets},
BlogCatalog and Flickr\footnote{\url{http://people.tamu.edu/~xhuang/Code.html}} are social networks.
The posted keywords or tags in BlogCatalog and Flickr networks are used as node attribute information. There are three air-traffic networks \cite{ribeiro2017struc2vec}: Brazil, Europe and USA, where each node corresponds to an airport and edge indicates the existence of commercial flights between the airports. Their class labels are assigned based on the level of activity measured by flights or people that passed the airports. Data statistics are summarized in Table \ref{node_datasets}.
For those networks without node attributes, we use the one-hot encoding of node degrees. In BlogCatalog, Flickr and other air-traffic networks, node class labels are available. In Facebook and Wiki-Vote, we use the degree-induced class labels by labeling the node according to its degree value.

In addition, we use four bioinformatics networks to evaluate the model performance on graph classification, including MUTAG, PTC, PROTEINS and ENZYMES\footnote{\url{https://ls11-www.cs.tu-dortmund.de/staff/morris/graphkerneldatasets}} where the nodes are associated with categorical input features. The detailed statistics for these bioinformatics networks are summarized in Table \ref{graph_datasets}.

\noindent\textbf{Model Configuration: } We adopt two hidden layers followed by the softmax activation layer in \demonet, where the proposed multi-task feature learning schemes in Eq. (7) and (10) are applied to each hidden layer for neighborhood aggregation (termed as {\demonet (weight)} and {\demonet (hash)}, respectively). In addition, we apply Adam optimizer \cite{kingma2014adam} with the learning rate 0.005 on the cross-entropy loss to train our models. To prevent our models from over-fitting, we adopt the dropout \cite{srivastava2014dropout} with $p = 0.6$ and $L_2$ regularization with $\lambda=0.0005$. The hidden layer size of neural units is set as 64. An early stopping strategy with a patience of 100 epochs on validation set is applied in our experiments.

\noindent\textbf{Baseline Methods: } The baseline methods used in our experiments are given below: (1) node-level graph neural networks: GCN \cite{kipf2016semi}, GCN\_cheby \cite{kipf2016semi}, GraphSAGE (mean aggregator) \cite{hamilton2017inductive}, Union \cite{li2018deeper}, Intersection \cite{li2018deeper} and GAT \cite{velickovic2017graph}; (2) node-level structure-aware embedding approaches: RolX \cite{henderson2012rolx}, struc2vec \cite{ribeiro2017struc2vec} and GraphWAVE \cite{donnat2018learning}; (3) graph-level graph neural networks: DCNN \cite{atwood2016diffusion}, PATCHY-SAN \cite{niepert2016learning} and DIFFPOOL \cite{ying2018hierarchical}; (4) deep graph kernel: DeepWL \cite{yanardag2015deep}. In our experiments, all the baseline models used the default hyperparameters suggested in the original papers.

All our experiments are performed on a Windows machine with four 3.60GHz Intel Cores and 32GB RAM. The source code will be available at \url{https://github.com/jwu4sml/DEMO-Net}.

\begin{table}
\caption{Data sets for node classification}
\label{node_datasets}
\small
\begin{tabular}{|l|c|c|c|c|}
\hline
\multicolumn{1}{|c|}{Data sets} & \# nodes & \# edges & \# classes & \# attributes \\\hline
Facebook & 4039 & 88234 & 4 & - \\ \hline
Wiki-Vote & 7115 & 103689 & 4 & - \\ \hline
BlogCatalog                   & 5196    & 171743  & 6         & 8189         \\\hline
Flickr                        & 7575    & 239738  & 9         & 12047        \\\hline
Brazil                        & 131     & 1038    & 4         & -            \\\hline
Europe                        & 399     & 5995   & 4         & -            \\\hline
USA                           & 1190    & 13599   & 4         & -           \\ \hline
\end{tabular}
\vspace{-3mm}
\end{table}

\begin{table}
\caption{Data sets for graph classification}
\label{graph_datasets}
\small
\begin{tabular}{|l|c|c|c|c|}
\hline
\multicolumn{1}{|c|}{Data sets} & \# graphs & \# classes & Avg \# nodes & \# attributes \\ \hline
MUTAG                           & 188      & 2         & 17.9        & 7            \\ \hline
PTC                             & 344      & 2         & 25.5        & 19           \\ \hline
PROTEINS                        & 1113     & 2         & 39.1        & 3            \\ \hline
ENZYMES                         & 600      & 6         & 32.6        & 3            \\ \hline
\end{tabular}
\vspace{-3mm}

\end{table}

\subsection{Node Classification}
For a fair comparison of different architectures \cite{shchur2018pitfalls}, we use different train/validation/test splits of the networks on node classification.
For social networks, we randomly choose 10\% and 20\% of the graph nodes as the training and validation set, respectively, and the rest as the test set. For air-traffic networks, the training, validation and test sets are randomly assigned with equal number of nodes. We run 10 times and report the mean accuracy with the standard variance for performance comparison. As shown in Table \ref{tab:node_classification}, we report the classification results on the real networks where the best results are indicated in bold. It can be observed that the proposed \demonet models significantly outperform other graph neural networks (answering \textbf{Q1}). In particular, our \demonet\ models are at least 10\% higher on mean accuracy over baseline methods. One explanation is that baseline methods focus on preserving the node proximity by roughly mixing a node with its neighbors, whereas our proposed \demonet\ models capture the degree-specific structure to distinguish the structural roles of nodes in the networks.

\begin{figure}[t]
\includegraphics[width = 3.35in]{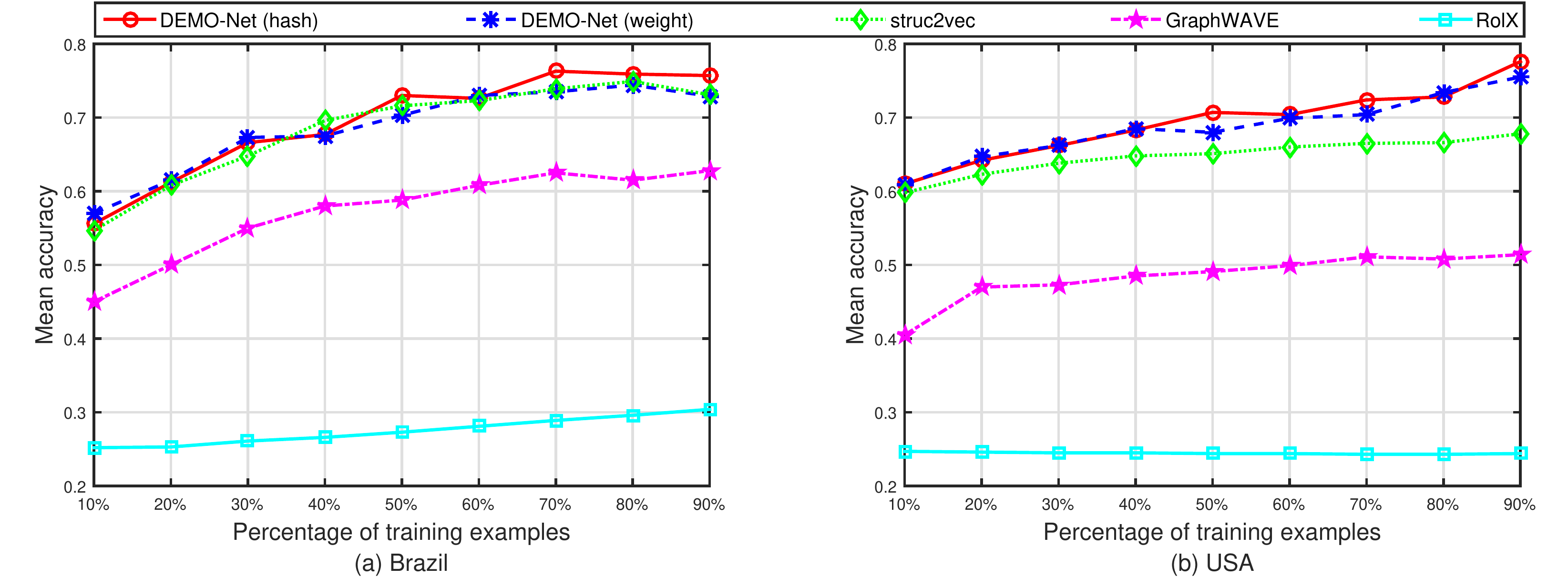}
\caption{Node-level classification accuracy on the Brazil and USA air-traffic networks using different train/test splits}
\label{struc_indentity}
\vspace{-4mm}
\end{figure}

We also evaluate the performance of our models against three structure-aware embedding approaches: RolX, struc2vec and GraphWAVE. All of them are unsupervised embedding approaches identifying the structural roles of nodes in the networks. Following \cite{ribeiro2017struc2vec}, we use the one-vs-rest logistic regression with L2 regularization to train a classifier for node representations learned by baseline methods. Here we consider using different train-test splits where the percentage of training nodes ranges from 10\% to 90\% and the rest is used for testing. The experimental results on the Brazil and USA air-traffic networks are provided in Figure \ref{struc_indentity}. We observe that our proposed \demonet\ models outperform the comparison methods across all the data sets (answering \textbf{Q2}). Besides, the structure roles identified by those baselines only represent the local graph structure without considering node attributes. Instead, both topological information and node attributes are captured in our \demonet\ models when learning node representation.

\begin{table*}
\centering
\caption{Node-level classification accuracy (mean $\pm$ standard variance) on the social and air-traffic networks}
\begin{tabular}{|l|c|c|c|c|c|c|c|}
\hline
\multirow{2}{*}{}     & \multicolumn{4}{c|}{Social networks} & \multicolumn{3}{c|}{Air-traffic networks}   \\ \cline{2-8} 
                    & Facebook     &  Wiki-Vote & BlogCatalog       & Flickr           & Brazil          & Europe          & USA \\ \hline
GraphSAGE \cite{hamilton2017inductive} & 0.389 \textpm 0.019 & 0.245 \textpm 0.000 & 0.828 \textpm 0.007& 0.641 \textpm 0.006& 0.404 \textpm 0.035 & 0.272 \textpm 0.022& 0.316 \textpm 0.022\\ \hline
GCN~\cite{kipf2016semi}       & 0.575 \textpm 0.013 & 0.329 \textpm 0.029 & 0.720 \textpm 0.013   & 0.546 \textpm 0.019  & 0.432 \textpm 0.064 & 0.371 \textpm 0.046 &  0.432 \textpm 0.022   \\ \hline
GCN\_cheby~\cite{kipf2016semi}    &     0.646 \textpm 0.012        & 0.495 \textpm 0.016      & 0.686 \textpm 0.037   &   0.479 \textpm 0.023               & 0.516 \textpm 0.070 & 0.460 \textpm 0.038 &  0.526 \textpm 0.045   \\ \hline
{Union}~\cite{li2018deeper}    & 0.600 \textpm 0.000       &  0.463 \textpm 0.000   & 0.730 \textpm 0.000   & 0.566 \textpm 0.000  & 0.466 \textpm 0.006 & 0.418 \textpm 0.002 & 0.582 \textpm 0.000    \\ \hline
{Intersection}~\cite{li2018deeper}  & 0.598 \textpm 0.000 & 0.462 \textpm 0.000     & 0.725 \textpm 0.000   & 0.557 \textpm 0.000  & 0.459 \textpm 0.003 & 0.443 \textpm 0.002 & 0.573 \textpm 0.000    \\ \hline
GAT~\cite{velickovic2017graph}      &   0.570 \textpm 0.036      &     0.594 \textpm 0.070              &   0.663 \textpm 0.000                &   0.359 \textpm 0.000               & 0.382 \textpm 0.126 & 0.424 \textpm 0.073 &  0.585 \textpm 0.021   \\ \hline
\hline
\demonet (hash)                 &  0.887 \textpm 0.020    &  0.997 \textpm 0.000     & 0.849 \textpm 0.006                  &    \textbf{0.678 \textpm 0.010}              & \textbf{0.614 \textpm 0.069} & \textbf{0.479 \textpm 0.064} &  \textbf{0.659 \textpm 0.020}   \\ \hline
\demonet (weight)   & \textbf{0.919 \textpm 0.003} & \textbf{0.998 \textpm 0.000} &    \textbf{0.849 \textpm 0.000}   & 0.656 \textpm 0.000  & 0.543 \textpm 0.034 & 0.459 \textpm 0.025 & 0.647 \textpm 0.021    \\ \hline
\end{tabular}
\label{tab:node_classification}
\end{table*}

\subsection{Graph Classification}
We use four public graph classification benchmarks to evaluate the proposed \demonet\ models with the degree-specific graph-level pooling scheme. DCNN \cite{atwood2016diffusion}, PATCHY-SAN \cite{niepert2016learning} and DIFFPOOL \cite{ying2018hierarchical} adopted the end-to-end training architectures for supervised graph classification. For unsupervised graph kernel method DeepWL \cite{yanardag2015deep}, we use the one-vs-rest logistic regression with L2 regularization to train a supervised classifier for graph classification. We also consider our model variants (denoted as \demonet\_m(hash) and \demonet\_m(weight) respectively) which replace the proposed degree-specific graph-level pooling with mean-pooling scheme \cite{atwood2016diffusion}. The input graphs are randomly assigned to the training, validation, or test set where each set has the same number of nodes.

The graph classification results are shown in Table \ref{tab:graph_classification} where the best results are indicated in bold. It is observed that (1) compared to the existing mean-pooling method, the proposed degree-specific pooling method improves the model performance in most cases, which is consistent with our analysis in Section 4.2; (2) the classification results of our \demonet\ models are comparable to other graph neural networks and graph kernel method (answering \textbf{Q3}). Moreover, on MUTAG and ENZYMES data sets, our proposed \demonet (weight) outperforms the baseline methods. One explanation might be that the graph representation generated by \demonet\ explicitly preserves the degree-specific graph structure information.

\begin{table}
    \centering
    \caption{Graph-level classification on the real networks}
    \small
    \begin{tabular}{|l|c|c|c|c|}
    \hline
         &  MUTAG & PTC  & PROTEINS & ENZYMES   \\ \hline
         DeepWL~\cite{yanardag2015deep} & 0.733 & 0.537  & 0.680  & 0.210   \\ \hline
         DCNN~\cite{atwood2016diffusion}   & 0.670  & 0.572 & 0.579   & 0.160  \\ \hline
         PATCHY-SAN~\cite{niepert2016learning} & 0.795  & 0.568  & 0.714  & 0.170   \\ \hline
         DIFFPOOL~\cite{ying2018hierarchical} & 0.663  & 0.251 & \textbf{0.733 } & 0.184  \\ \hline
         \hline
         \demonet\_m(hash)   & 0.760  & \textbf{0.586}  & 0.617  & 0.236  \\ \hline
         \demonet\_m(weight)   & 0.798  & 0.550  & 0.616 &  0.251 \\ \hline
         \demonet (hash)   & 0.771  & 0.563  & 0.705 & 0.251  \\ \hline
         \demonet (weight) & \textbf{0.814} & 0.572 & 0.708  & \textbf{0.272} \\ \hline
    \end{tabular}
    \label{tab:graph_classification}
    \vspace{-3mm}
\end{table}

% \subsection{Case Study: Structural Identity Analysis}
% \textbf{Synthetic "cycle" network: }

% \noindent\textbf{Mirrored Karate network: }

\subsection{Efficiency Analysis}
It is easy to show that the time complexity of each layer in our proposed \demonet (hash) model is $O(nFF'+TF+nHF'+mF')$ where $n$ and $m$ are the number of nodes and edges in the graph, respectively, $F$ and $F'$ are the dimensionalities of input and output features at each layer, respectively, $T$ is the number of tasks (degree values) in the graph, and $H$ is the hashing dimension. By observing that $T \leq n$ in the networks, its time complexity would be $O(nFF'+mF')$, which is on par with GCN and GAT models. Similarly, we can show that the time complexity of each layer in \demonet(weight) is $O(T(nFF'+mF'))$. When $T \ll n$ and $T \ll m$, it also scales linearly with respect to the number of nodes and edges.

\begin{figure}
\includegraphics[width = 3.35in]{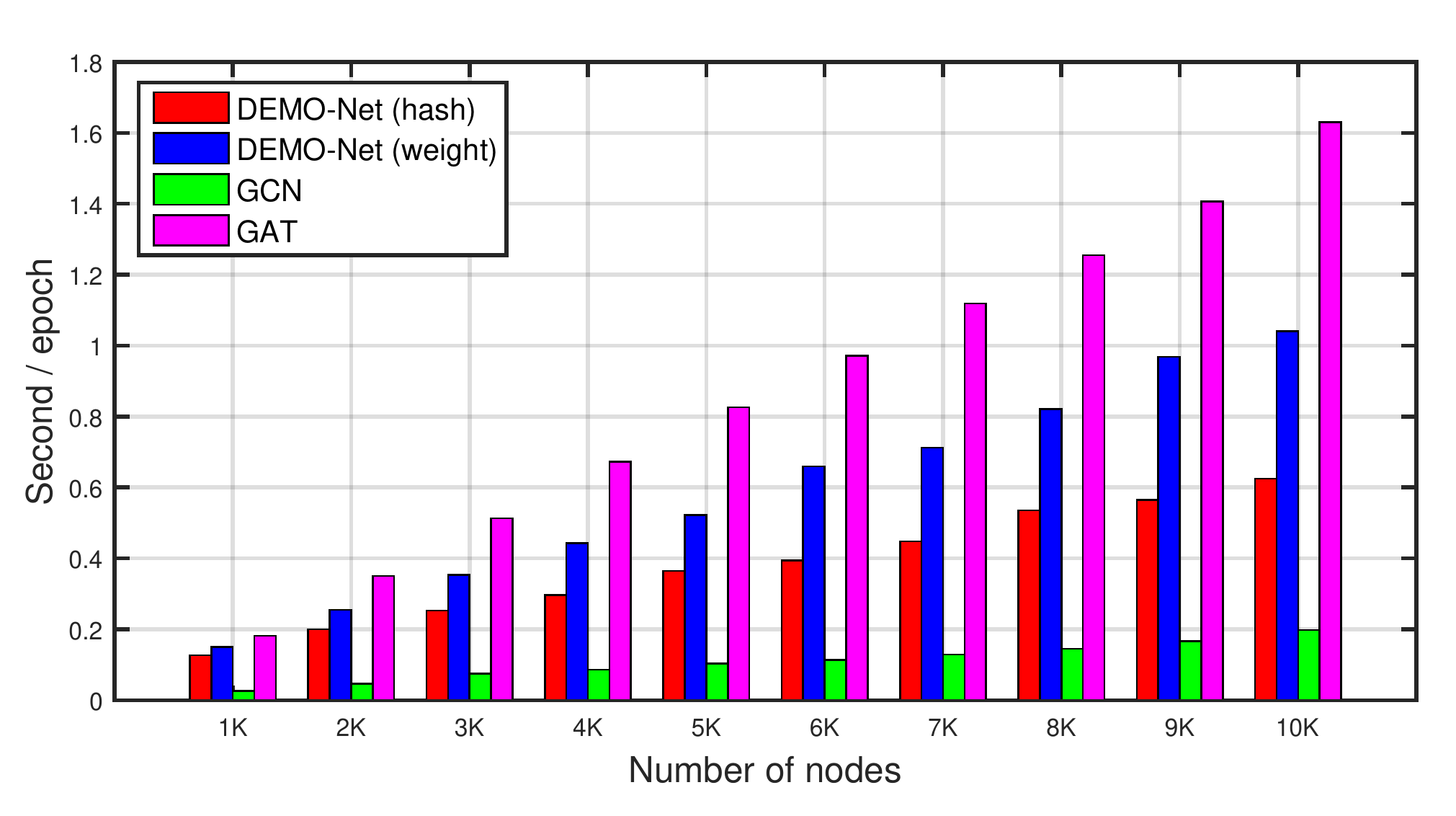}
\caption{Running time per epoch (best seen in color)}
\label{running_time}
\vspace{-3mm}
\end{figure}

Following \cite{kipf2016semi}, we report the running time (measured in seconds wall-clock time) per epoch (including forward pass, cross-entropy calculation, backward pass) on a synthetic network assigning $2n$ edges uniformly at random. As shown in Figure \ref{running_time}, we observe that (answering \textbf{Q4}) (1) the wall-clock time of our proposed \demonet\ model is linear with respect to the number of nodes; (2) our models are much more efficient than GAT on node classification task.

\section{Conclusions}
In this paper, we focus on building a degree-specific graph neural network for both node and graph classification. We start by analyzing the limitations of the existing graph neural networks from the perspective of Weisfeiler-Lehman graph isomorphism test. Furthermore, it is observed that the graph convolution should have the following properties: {\em seed-oriented}, {\em degree-aware}, {\em order-free}. To this end, we propose a generic graph neural network model named \demonet\, which formulates the feature aggregation into a multi-task learning problem according to nodes' degree values. In addition, we also present a novel graph-level pooling method for learning graph representations provably lying in a degree-specific Hilbert kernel space. The extensive experiments on real networks demonstrate the effectiveness of our \demonet\ algorithm.

\begin{acks}
This work is supported by the United States Air Force and DARPA under contract number FA8750-17-C-0153, National Science Foundation under Grant No. IIS-1552654, Grant No.
IIS-1813464 and Grant No. CNS-1629888, the U.S. Department of Homeland Security
under Grant Award Number 17STQAC00001-02-00, and an IBM Faculty Award. The views and conclusions are those of the authors and should not be interpreted as representing
the official policies of the funding agencies or the government.
\end{acks}

%% file: 008appendix.tex
\appendix
\section{Appendix for Reproducibility}
To better reproduce the experimental results, we provide additional details about the algorithms.

{\bf Proof of Theorem~\ref{T: existence}}.
Theorem~\ref{T: existence} says that there exist mapping functions $f_s$ and $\{ f_{deg} | deg \in degree(G) \}$ such that for any two subtrees in $\mathcal{T}$, the function $f: \mathcal{T} \rightarrow \mathbb{R}^d$ defined in Eq. (5) maps them to different feature vectors if they are not structurally identical.
\begin{proof}
Let $\mathcal{T}_s =\{h_v | v \in V\}$ denote the seed set in $\mathcal{T}$ and $d_{m} = max\{deg\} + 1$ the maximum degree values plus one. Becuase $\mathcal{T}$ is countable, there exists an injective function $Z: \mathcal{T} \rightarrow \mathbb{N}$ that maps each subtree from $\mathcal{T}$ to an unique natural number. It can be observed that $\mathbb{N}$ can be divided into $d_m$ disjoint sets: $\mathbb{N}_0 = \{i*d_m\}_{i=0}^{\infty}$, $\mathbb{N}_1 = \{i*d_m + 1\}_{i=0}^{\infty}$, $\cdots$, $\mathbb{N}_{d_m-1} = \{i*d_m + d_m - 1\}_{i=0}^{\infty}$.

There exists an injective function $Z_s: \mathcal{T}_s \rightarrow \mathbb{N}_0$ that maps each seed from $\mathcal{T}_s$ to an unique natural number in $\mathbb{N}_0$. Let $\mathcal{T}_{i} = \{h_{N(v)} | v \in V~ \text{and}~ deg(v)=i\}$ denote the neighbor set consisting of the seeds' neighbors when their degree values are equal to $i$. Because $\mathcal{T}$ is countable, all the subsets $\mathcal{T}_{i}~(1\leq i \leq max\{deg\})$ are countable. There exists the injective, symmetric function $Z_i: \mathcal{T}_i \rightarrow \mathbb{N}_i$ that maps each element from $\mathcal{T}_i$ to an unique real number in $\mathbb{N}_i$. Moreover, there is a function $Z_f: \mathcal{T} \rightarrow \mathbb{N}^2$ that maps each subtree from $\mathcal{T}$ to an unique feature vector in $\mathbb{N}^2$ when $Z_f(h_v, h_{N(v)}) = Z_s(h_v) \circ Z_{deg(v)}(h_{N(v)})$. Please note that the structurally identical subtrees would be considered the same one when the degree-specific function $Z_i$ is symmetric.

It is easy to construct an injective function $g: \mathbb{N}^2 \rightarrow \mathbb{R}^d$. Based on the properties of injective function, $g(Z_f(\cdot))$ will be injective function that maps any two subtrees in $\mathcal{T}$ to different feature vectors in $\mathbb{R}^d$ if they are not structurally identical, which completes the proof.
\end{proof}

\noindent{\bf Proof of Theorem~\ref{T: kernel}}.
Theorem~\ref{T: kernel} says that the graph representation $h_G$ learned in Eq. (12) belongs to the Reproducing Kernel Hilbert Space (RKHS) of kernel $\mathcal{K}_{\sigma,DWL}(\cdot,\cdot)$.
\begin{proof}
Let $h_{G_k}[i] = \sum\nolimits_{v \in V}h_v^k \cdot \delta(deg(v), d_i)$ denote the feature vector of graph $G_k$ for nodes with degree value $d_i$. Let $h_G[k, i, j] = h_{G_k}[i][j]$ denote the $j^{\textrm{th}}$ element of $h_{G_k}[i]$. Our graph convolution (feature aggregation) function can be written as:
\begin{equation}
    h_v^k = \sigma(W_0^k h_v^{k-1} \circ \sum\nolimits_{u\in N(v)} \hat{W}^k_{deg(v)}h_u^{k-1})
\end{equation}
where $\hat{W}^k_{deg(v)}$ represents the degree-specific parameters, and more specifically, $\hat{W}_{deg(v)}=W_g^k + W_{deg(v)}^k$ for degree-specific weight matrix in Eq. (7) and $\hat{W}_{deg(v)}=W^k(\phi_g(\cdot) + \phi_{deg}(\cdot))$. Because we use the concatenation operator $\circ$ to combine the learned features of seed and its neighborhood, it holds that $h_{G_k}[i][j]$ lies in \textbf{either} seed's feature $\sigma(W_0^k h_v^{k-1})$ \textbf{or} $\sigma(\sum\nolimits_{u\in N(v)} \hat{W}^k_{deg(v)}h_u^{k-1})$, but not both.

Let $w_{0j}$ denote the $j^{\textrm{th}}$ row from $W_0^k$. To show our results, we construct a $k$-regular "reference graph" $G_{r_k} = (V_{r_k}, E_{r_k})$ which has the same nodes as the input graph $G$ (i.e., $V_{r_k} = V$). Its degree value $k$ is $d_i$ and each node in "reference graph" is associated with the same feature vector $w_{0j}/n$. Then when $h_{G_k}[i][j]$ lies in the seed's feature $\sigma(W_0^k h_v^{k-1})$, we have:
\begin{equation}
    \begin{aligned}
        & h_G[k, i, j] = h_{G_k}[i][j] \\
        =& \sum\nolimits_{v \in V} h_v^k[j] \cdot \delta(deg(v), d_i) \\
        =& \sigma \Big(\sum\nolimits_{v\in V}  \delta(deg(v), d_i) \cdot \left\langle{h_v^{k-1}, {w}_{0j}}\right\rangle \Big) \\
        =& \sigma \Big(\frac{1}{n} \sum\nolimits_{v\in V} \sum\nolimits_{r\in V_{r_k}} \delta(deg(v), deg(r)) \cdot \left\langle{h_v^{k-1}, {w}_{0j}}\right\rangle \Big) \\
        = & \sigma\Big(\mathcal{K}_{DWL}({G_{k-1}}, G_{r_k}) \Big)
    \end{aligned}
\end{equation}

The lemma 1 in \cite{lei2017deriving} holds that for activation functions $\sigma$, there exists kernel functions $\mathcal{K}_{\sigma}(\cdot,\cdot)$ and the underlying mapping $\phi_{\sigma}(\cdot)$ such that $f(x)=\sigma(w^T x)= \left\langle{\phi_{\sigma}(x), \psi(w)}\right\rangle$ for some mapping function $\psi(w)$ constructed from $w$. Therefore, we have:
\begin{equation}
    h_G[k, i, j]=\mathcal{K}_{\sigma,DWL}({G_{k-1}}, G_{\sigma, r_k})
\end{equation}
where $\mathcal{K}_{\sigma,DWL}(\cdot, \cdot)$ is the composition of $\mathcal{K}_{\sigma}(\cdot,\cdot)$ and $\mathcal{K}_{DWL}(\cdot,\cdot)$, and $\mathcal{K}_{\sigma,DWL}(x, y) = \phi_{\sigma}(\phi_{DWL}(x))^T \phi_{\sigma}(\phi_{DWL}(y))$. And $G_{\sigma, r_k}$ is the "reference graph" constructed from model parameters and activation function.

Let $\hat{w}_{d_ij}$ denote the $j^{\textrm{th}}$ row from $\hat{W}_{deg(v)}^k$ with $deg(v)=d_i$. Similarly, we construct a $k$-regular "reference graph" $\hat{G}_{r_k} = (\hat{V}_{r_k}, \hat{E}_{r_k})$ which has the same nodes as the input graph $G$ with degree value $d_i$. Each node in this "reference graph" is associated with the same feature vector $\hat{w}_{d_ij}/n$.
when when $h_{G_k}[i][j]$ lies in the neighborhood's feature $\sigma(\sum\nolimits_{u\in N(v)} \hat{W}^k_{deg(v)}h_u^{k-1})$, we have:
\begin{equation}
    \begin{aligned}
        & h_G[k, i, j] = h_{G_k}[i][j] \\
        =& \sum\nolimits_{v \in V} h_v^k[j] \cdot \delta(deg(v), d_i) \\
        =&\sigma\Big( \sum\nolimits_{v \in V} \sum\nolimits_{u\in N(v)} \delta(deg(v), d_i) \cdot \left\langle{h_u^{k-1}, \hat{w}_{d_ij}}\right\rangle \Big) \\
        =& \sigma\Big( \frac{1}{n} \sum_{v\in V} \sum_{r\in \hat{V}_{r_k}} \delta(deg(v), deg(r)) \cdot \left\langle{\sum_{u \in N(v)} h_u^{k-1}, \hat{w}_{d_ij}}\right\rangle \Big) \\
        =& \sigma\Big(\mathcal{K}_{DWL}({G_{k-1}}, \hat{G}_{r_k}) \Big)
    \end{aligned}
\end{equation}
Please notice that in this case, node features are assumed to be the sum of neighborhood features. And moreover, it can be written as:
\begin{equation}
    h_G[k, i, j]=\mathcal{K}_{\sigma,DWL}({G_{k-1}}, \hat{G}_{\sigma, r_k})
\end{equation}
where $\hat{G}_{\sigma, r_k}$ is the "reference graph" constructed from model parameters and activation function.
Therefore, the graph representation $h_G$ belongs to the RKHS of kernel $\mathcal{K}_{\sigma,DWL}(\cdot,\cdot)$, which completes the proof.
\end{proof}

% \subsection{Implementation Details}

% \subsection{Additional Experimental Results}
% Here, we provide some additional experimental results on parameter sensitivity. 

% \noindent {\bf Parameter Sensitivity}. We present the parameter study of the proposed \demonet\ models on both node and graph classification. Figure \ref{param_layers} shows the classification results on Facebook and MUTAG data sets. It can be observed that the model performance on node classification would be improved when the number of layers becomes larger, and the improvement becomes marginal when it is larger than 4. That is also consistent with the existing work on graph neural network \cite{kipf2016semi, atwood2016diffusion, li2018deeper}. On MUTAG data set, we obtain the highest mean accuracy when the number of layers is 2. The deeper model would not improve the graph classification performance due to the small average number of nodes. In order to balance the effectiveness and efficiency, we choose to use two hidden layers in the proposed \demonet\ models for all the experiments.

% \begin{figure}
% \includegraphics[width = 3.35in]{figures/param_layers.pdf}
% \caption{Impact of the number of hidden layers on: (a) Facebook, (b) MUTAG}
% \label{param_layers}
% \end{figure}